\documentclass{article}

\usepackage{microtype}
\usepackage{graphicx}
\usepackage{booktabs} % for professional tables

\usepackage{hyperref}
\usepackage{multirow}
\usepackage{times}
\usepackage{epsfig}
\usepackage{graphicx}
\usepackage{chngpage}
\usepackage{caption}
\usepackage[table,xcdraw]{xcolor}
\usepackage{listings}
\usepackage{subcaption}
\usepackage{sidecap}
\usepackage{placeins}

\usepackage[utf8]{inputenc} % allow utf-8 input
\usepackage[T1]{fontenc}    % use 8-bit T1 fonts
\usepackage{hyperref}       % hyperlinks
\usepackage{url}            % simple URL typesetting
\usepackage{booktabs}       % professional-quality tables
\usepackage{amsfonts}       % blackboard math symbols
\usepackage{nicefrac}       % compact symbols for 1/2, etc.
\usepackage{microtype}      % microtypography
\usepackage{xcolor}         % colors
\usepackage{sidecap}
\usepackage{xspace}
\usepackage{tabularx}

\newcommand{\fedavg}{\texttt{FedAvg}\xspace}
\newcommand{\fedfwd}{\texttt{FedFwd}\xspace}
\newcommand{\Symba}{\texttt{SymBa}\xspace}

\usepackage[accepted]{icml2023}

\usepackage{amsmath}
\usepackage{amssymb}
\usepackage{mathtools}
\usepackage{amsthm}

\usepackage[capitalize,noabbrev]{cleveref}

\theoremstyle{plain}

\theoremstyle{definition}

\theoremstyle{remark}

\usepackage[textsize=tiny]{todonotes}

\icmltitlerunning{Federated Learning without Backpropagation}

\begin{document}

\twocolumn[
\icmltitle{\fedfwd: Federated Learning without Backpropagation}
% Exploring Forward-Forward Propagation: A Potential Alternative to Backpropagation in Federated Learning



\icmlsetsymbol{equal}{*}

\begin{icmlauthorlist}
\icmlauthor{Seonghwan Park}{yyy}
\icmlauthor{Dahun Shin}{yyy}
\icmlauthor{Jinseok Chung}{yyy}
\icmlauthor{Namhoon Lee}{yyy}
\end{icmlauthorlist}

\icmlaffiliation{yyy}{POSTECH, Pohang, Republic of Korea}

\icmlcorrespondingauthor{Seonghwan Park}{shpark97@postech.ac.kr}
\icmlcorrespondingauthor{Dahun Shin}{dahunshin@postech.ac.kr}
\icmlcorrespondingauthor{Jinseok Chung}{jinseokchung@postech.ac.kr}
\icmlcorrespondingauthor{Namhoon Lee}{namhoonlee@postech.ac.kr}

\icmlkeywords{Machine Learning, ICML}

\vskip 0.3in
]



\printAffiliationsAndNotice{}  % leave blank if no need to mention equal contribution

\begin{abstract}
In federated learning (FL), clients with limited resources can disrupt the training efficiency.
A potential solution to this problem is to leverage a new learning procedure that does not rely on backpropagation (BP).
We present a novel approach to FL called \fedfwd that employs a recent BP-free method by \citet{hinton2022forwardforward}, namely the Forward Forward algorithm, in the local training process.
\fedfwd can reduce a significant amount of computations for updating parameters by performing layer-wise local updates, and therefore, there is no need to store all intermediate activation values during training.
We conduct various experiments to evaluate \fedfwd on standard datasets including MNIST and CIFAR-10, and show that it works competitively to other BP-dependent FL methods.
\end{abstract}
\section{Introduction}
\label{submission}

Federated learning (FL) is a machine learning strategy that trains a global model on multiple local devices without sharing the data between them \citep{DBLP:journals/corr/McMahanMRA16}.
When applied to training large neural networks, however, FL is immediately challenged since the local clients are often restricted to permit a limited amount of compute and memory resources \citep{DBLP:journals/corr/abs-1912-04977}.
This could fundamentally restrict deploying large yet high-performing deep models.
Many studies suggest different ways to improve the efficiency of FL \citep{DBLP:journals/corr/Alistarh0TV16, stich2018local, DBLP:journals/corr/abs-1909-13014, NEURIPS2020_18df51b9}, and yet, they are mainly focused on reducing communication overhead rather than the local training cost.

In this work, we approach this problem quite differently by adopting a new learning procedure.
Precisely, we propose to replace the backpropagation procedure (BP) \citep{rumelhart1986learning} for computing gradients on local clients with the forward-forward algorithm (FF) that is recently introduced by \citet{hinton2022forwardforward}.
The resulting method--\fedfwd--replaces the forward and backward passes of BP by two forward passes to compute gradients, which can reduce the computational burden on local clients by eliminating the need to store all intermediate activations in memory.

We evaluate \fedfwd for MLP-like architectures on MNIST and CIFAR-10 image classification datasets and compare with the standard BP-based FL algorithm, i.e., \fedavg \citep{DBLP:journals/corr/McMahanMRA16} in terms of the test accuracy and training time.
As a result, we achieve $96.78\%$ test accuracy when training a $3$-layer MLP model on MNIST with an i.i.d setting, which is on a par with \fedavg.
We also show that \fedfwd can finish the training in the similar wall clock time to \fedavg without any optimized implementation. 
We demonstrate that a further modification to the objective function can stabilize training and yield faster convergence.

\section{Background}
\label{sec:relatedworks}
\subsection{Federated Learning (FL)}
As smartphones and wearable devices become increasingly powerful and more widely used, an intriguing learning paradigm has emerged. 
This paradigm sees deep learning models trained locally on distributed devices, rather than concentrating data within a single data center. 
This innovative approach is termed Federated Learning (FL) \citep{DBLP:journals/corr/McMahanMRA16}, and it has sparked numerous investigations into various issues such as privacy, robustness, and heterogeneity, leading to significant advancements.
However, no study has yet completely addressed the fundamental constraints of FL, namely, the limited memory and learning time. 
Various strategies have been proposed in an attempt to overcome these issues. 
Some approaches, from an optimization perspective, have aimed to calculate and accelerate convergence rates \citep{li2020convergence, yuan2021federated, li2020acceleration, chen2019federated}, while others have employed compression strategies to personalize the model and distribute only a part of it \citep{DBLP:journals/corr/abs-2012-04221, DBLP:journals/corr/abs-2002-07948, DBLP:journals/corr/abs-1912-00818}, or to learn a smaller model through knowledge distillation \citep{DBLP:journals/corr/abs-1910-03581, DBLP:journals/corr/abs-2105-10056}.
Although these methods have collectively improved FL and enhanced its practicality, none has provided an optimal solution. 
In this paper, we introduce an innovative algorithm that modifies the learning procedure itself, presenting a new fascinating strategy to address the inherent challenges of FL.

\subsection{Backpropagation Algorithm}
\label{sec:relff}
Backpropagation (BP) \citep{rumelhart1986learning}, the cornerstone of training artificial neural networks, is used to calculate the gradient of a loss function with respect to all weights in the network, thus facilitating optimization via gradient descent.
Its versatility extends to networks with differentiable activation functions and loss functions, fitting various architectures.
Despite its proven efficacy, BP comes with certain computational and memory-related drawbacks. 
Its sequential nature impedes parallel computation across layers, contributing to significant computational overhead due to step-wise gradient calculation from the output layer back to the input layer. 
Furthermore, BP requires the storage of intermediate states for all nodes in the network during the forward pass for utilization during the backward pass. 
This requirement results in memory usage that scales linearly with the size and depth of the network. 
These inefficiencies, especially in the context of larger and deeper networks, pose critical challenges to the enhancement of network availability and scalability.

\subsection{Federated Averaging (\fedavg)}
Federated Averaging (\fedavg), as initially proposed by \citet{DBLP:journals/corr/McMahanMRA16}, has significantly advanced the field by reducing overall communication costs.
This approach involves distributing a global model to each client as the first step.
These clients then undertake numerous stochastic gradient descent (SGD) iterations during their local training, making use of their private data. 
After completing this local training, the clients send their locally trained models back to the central server, which then aggregated to the global learning objectives across the entire network.
% [필요하면 넣기] This methodology, which allows for a global model update after several local learning sessions, is remarkably effective in reducing overall communication costs.
More specifically, the main goal of \fedavg is to optimize the following equation:
\begin{equation}
\min_{w} f(w) = \frac{1}{m}\sum_{i=1}^m (F_i(w) \vcentcolon = \mathbb{E}_{x_i \sim \mathcal{D}_i}[f_i(w; x_i)]),
\end{equation}
where $m$ denotes the number of clients participating in a Federated Learning (FL) system, and $F_i$ signifies the objective function for the $i$-th client $c_i$.
Given training samples $x_n$ and corresponding labels $y_n$, it is assumed that local clients $c_i$ address a local empirical risk minimization problem within their distinct data distributions $\mathcal{D}_i$.
\section{Proposed Method: \fedfwd}
\label{sec:methods}
We propose a federated learning algorithm, \fedfwd, which follows the core steps of \fedavg, encompassing three primary stages: 
1) the selection of a client subset at each iteration, 2) execution of local parameter updates, and 3) subsequent aggregation of these updates at the server. 
However, \fedfwd introduces key alterations to the learning procedure, which conventional BP-based FL algorithms overlook, and these modifications have the potential to lighten both the time and device constraints for clients.

\begin{figure*}[t]
    \centering
    \subfloat[\small Hidden Layer = 2 (i.i.d.)]{{\includegraphics[width=0.24\textwidth]{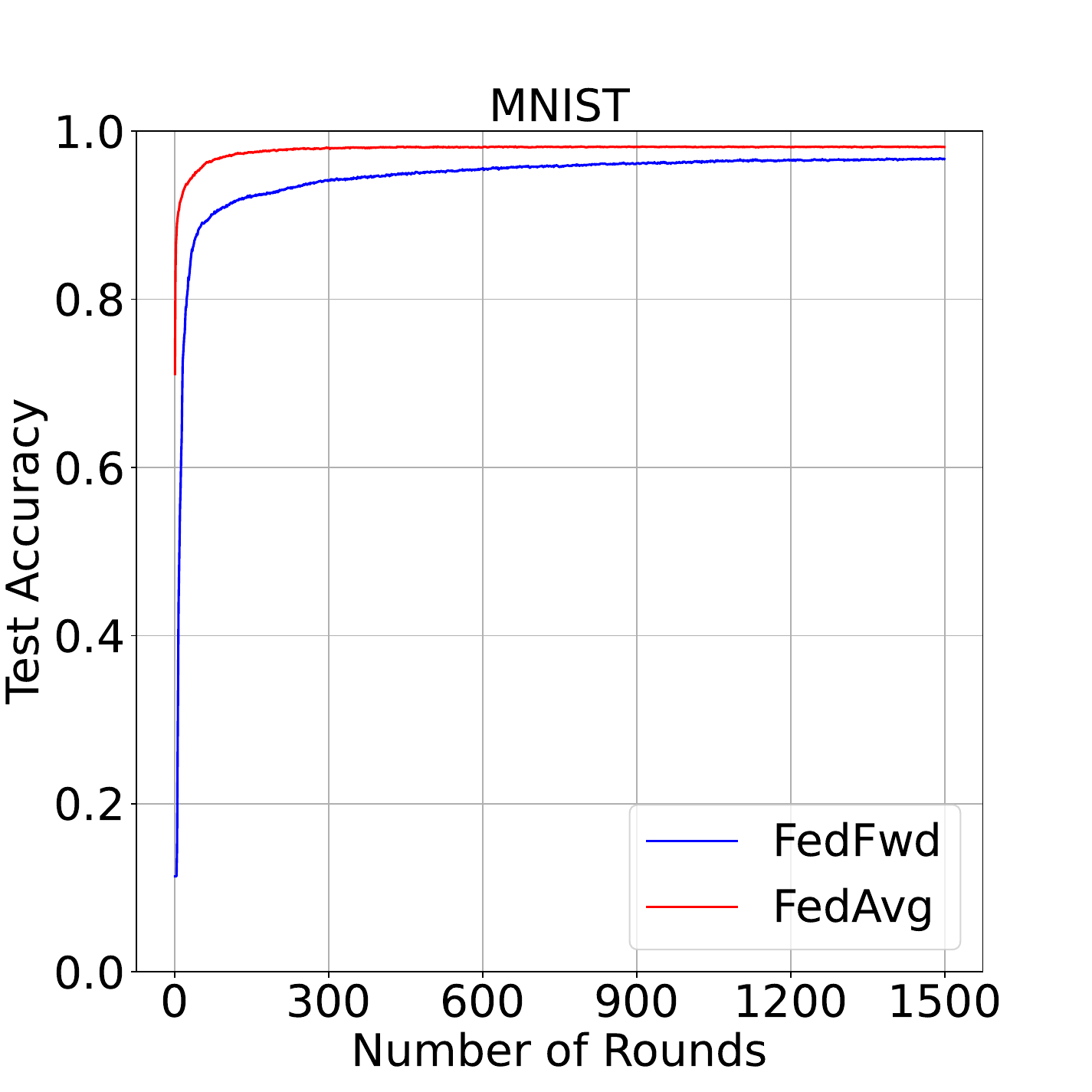} }}
    \subfloat[\small Hidden Layer = 3 (i.i.d.)]{{\includegraphics[width=0.24\textwidth]{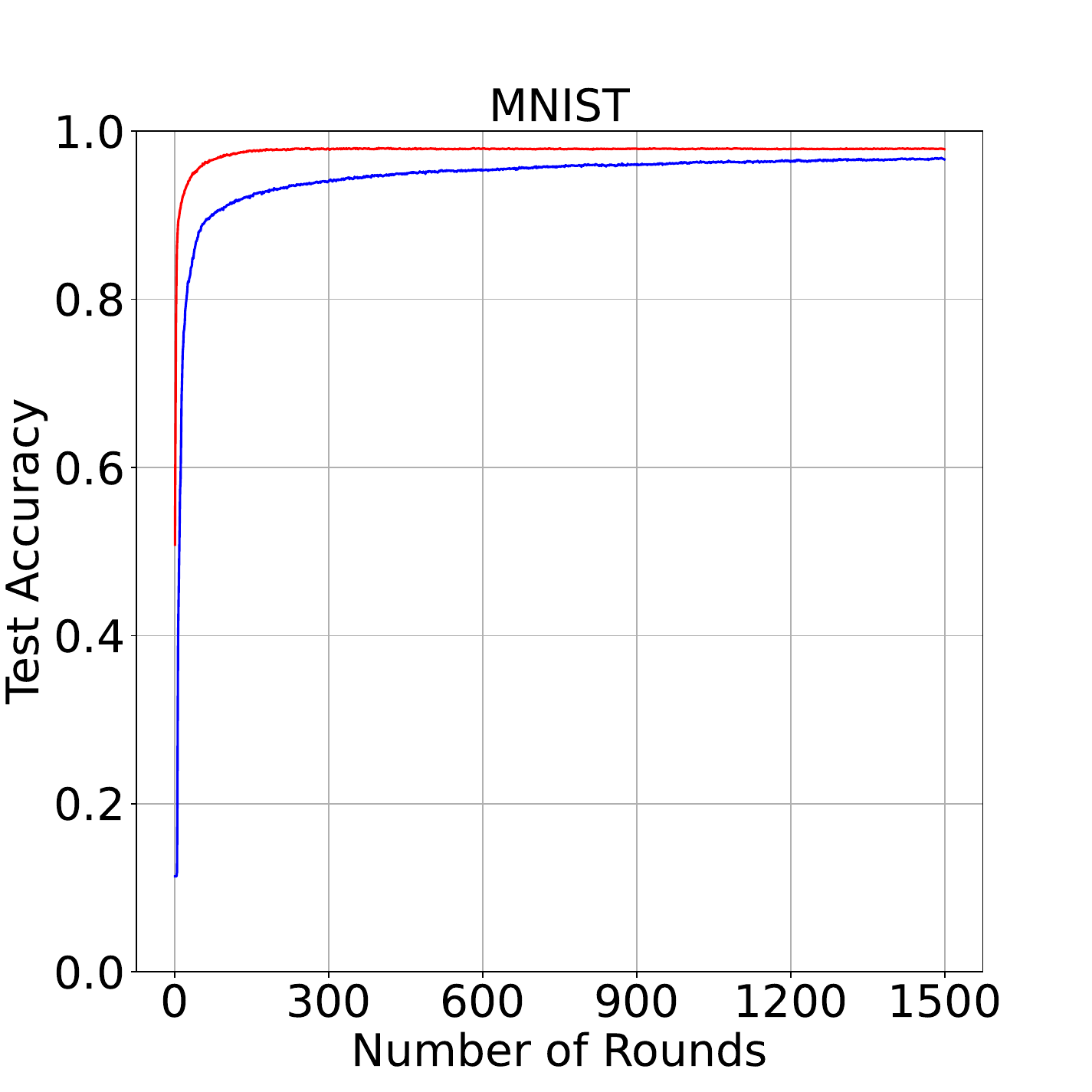} }}
    \subfloat[\small Hidden Layer = 2 (non-i.i.d.)]{{\includegraphics[width=0.24\textwidth]{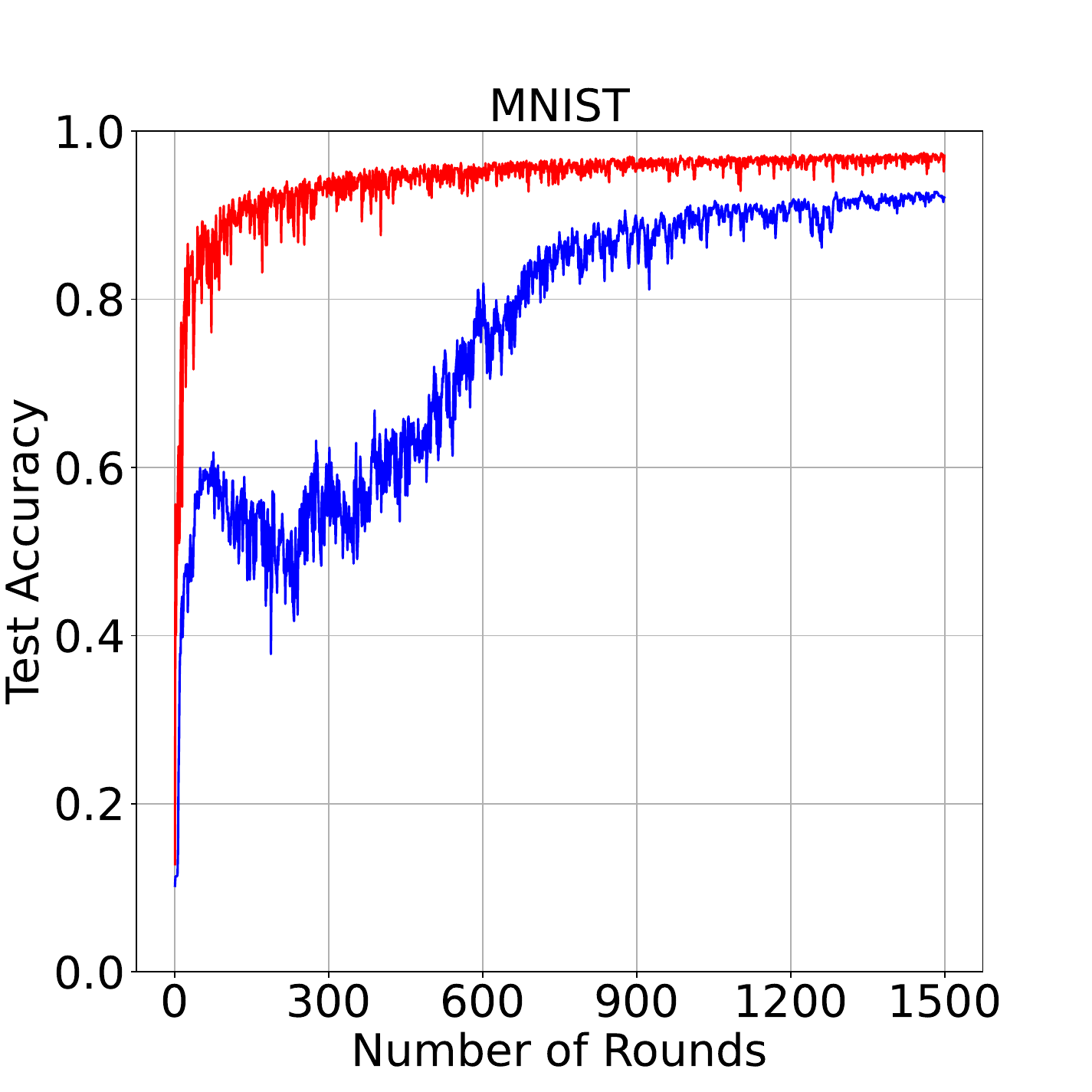} }}
    \subfloat[\small Hidden Layer = 3 (non-i.i.d.)]{{\includegraphics[width=0.24\textwidth]{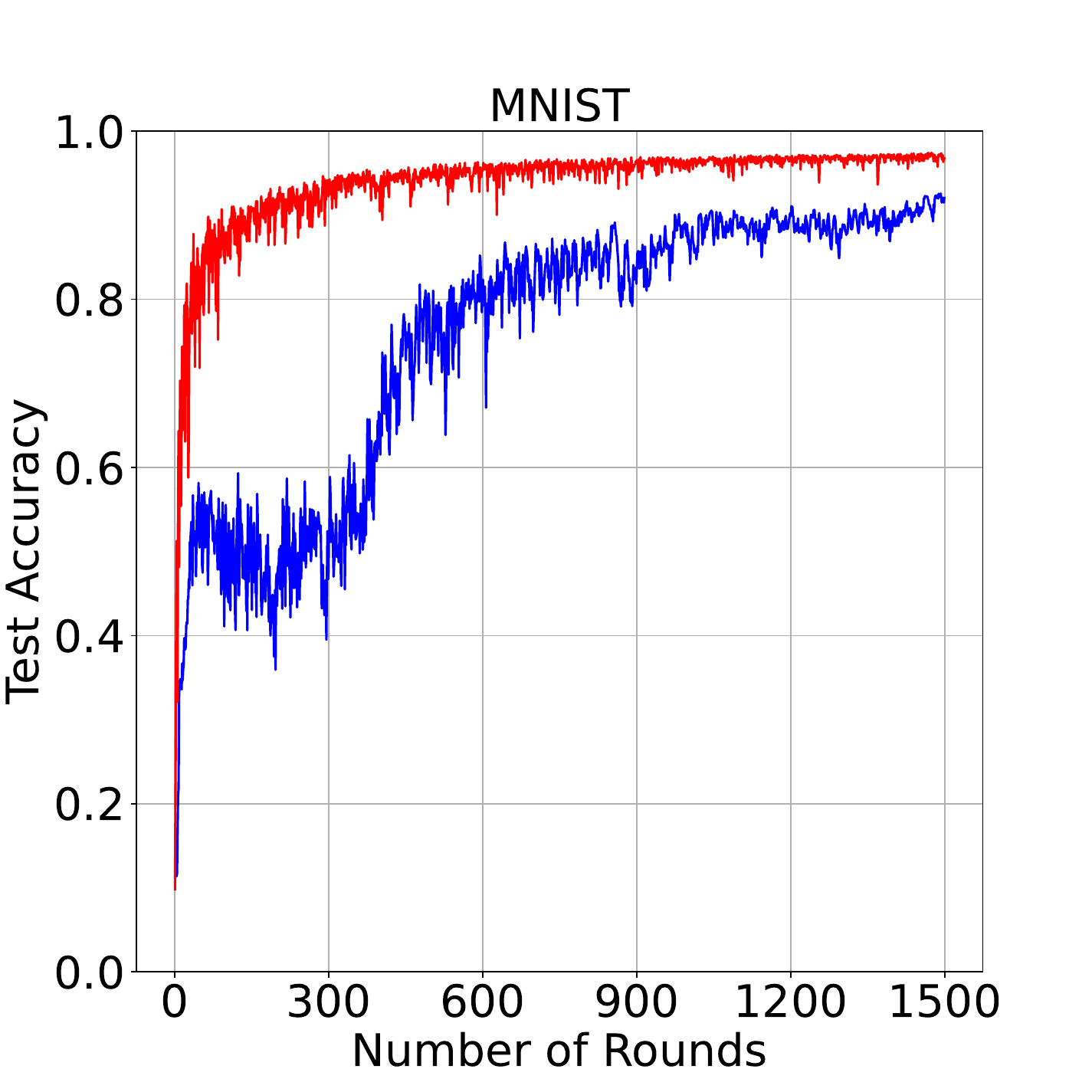} }}
    \captionsetup{width=.6\textwidth}
    \caption{
        \small \textbf{Comparison in performance and convergence speed between \fedfwd and \fedavg under both i.i.d. and non-i.i.d. settings.} 
        We alter the depth of the hidden layers to include both 2 and 3 layers. 
        The results of this experiment show that the \fedfwd algorithm demonstrates a performance and convergence rate similar to \fedavg under the i.i.d. setting. 
        However, under the non-i.i.d. setting, \fedfwd exhibits a slower convergence rate and performance compared to \fedavg.
    }
    \label{fig:learnable}
\end{figure*}

\textbf{Learning Procedure.}
The Forward-Forward algorithm, a greedy layer-wise learning technique, adopts an alternative approach to the traditional backpropagation's one forward and one backward pass by employing two forward passes. 
This algorithm uniquely trains each layer by leveraging a measure of goodness.
There are numerous potential ways to quantify goodness, but in this paper, we use the sum of the squared lengths of activity vectors. 
The goodness function is thus defined as:

\begin{equation}
goodness = \sum_j y_{j}^2,
\end{equation}

where $y_j$ represents the activity of hidden unit $j$. 
We assess the goodness of positive and negative samples separately. 
Each layer seeks to maximize the goodness of positive samples exceeding a threshold, $\theta$, while minimizing the goodness of negative samples beneath this threshold.
Given our usage of logistic functions and a threshold, the resulting objective functions take the form of probability functions:

\begin{equation}
\begin{cases*}
    p(positive) = \sigma (goodness_{x_{pos}} - \theta) \\
    p(negative) = \sigma (-goodness_{x_{neg}} + \theta)
\end{cases*}
\end{equation}

Positive data $x_{pos}$ refers to image data wherein some initial pixels have been replaced with the corresponding label's one-hot vector. 
In contrast, negative data $x_{neg}$ incorporates one-hot vectors of incorrect labels in place of some initial pixels.
Given that the data already incorporates one-hot vectors of labels, we employ gradient descent to maximize or minimize the probability function on a layer-by-layer basis.

Moreover, the \fedfwd algorithm applies layer normalization before passing the activity vector to the subsequent hidden layer. 
When the activities of a previous hidden layer are utilized as input for the next, it becomes relatively straightforward to distinguish between positive and negative samples by merely considering the length of the activity vector in the preceding hidden layer. 
Thus, there's no requirement for learning any new features. 
However, the layer normalization process eliminates all information previously used for the calculation of $\lq$goodness', compelling the subsequent hidden layer to learn new features from previously untapped information.
\section{Experiments}
In this section, we compare the test accuracy of \fedfwd and \fedavg by varying the size and depth of the hidden layers. We then evaluate the training speed of \fedfwd and \fedavg based on the size of the mini-batch. Additionally, we conduct supplementary experiments on \fedfwd with different objective functions. Detailed descriptions of the experimental procedures are provided in Appendix \ref{app:overallexp}.

\subsection{Experimental Details}
\textbf{Setup.}
To ensure a fair comparison, we design \fedfwd and \fedavg models to have the same number of layers and a similar number of parameters with a marginal parameter difference of approximately 1\%.
We conduct the tests under both i.i.d. and non-i.i.d. settings.
The former refers to a situation where the data is evenly distributed among the clients, while the latter refers to a situation where the data is not evenly distributed. 
We use the MNIST dataset, which is distributed across 100 clients.

\textbf{Hyperparameters.}
We set the number of selected clients to be 100, assuming that only 10\% of them participate in updating the global model. 
We conduct a local epoch, $E$, of 3 and 1500 global epochs. 
Each client has a local batch size of 10. 
Our learning rate is set to 0.003.

\subsection{Results}
\begin{table}[h]
\resizebox{\columnwidth}{!}{%
\begin{tabular}{|c|c|c|cl|cl|}
\hline
\rowcolor[HTML]{EFEFEF} 
\small \# Layers & \begin{tabular}[c]{@{}c@{}}\small Learning\\\small Procedure\end{tabular} & \small \# Params & \multicolumn{2}{c|}{\cellcolor[HTML]{EFEFEF}\begin{tabular}[c]{@{}c@{}}\small I.I.D.\\\small  Test Acc (\%)\end{tabular}} & \multicolumn{2}{c|}{\cellcolor[HTML]{EFEFEF}\begin{tabular}[c]{@{}c@{}}\small Non-I.I.D.\\\small  Test Acc (\%)\end{tabular}} \\ \hline
 & \small FedAvg (BP) &\small  0.64M & \multicolumn{2}{c|}{\small $98.19$} & \multicolumn{2}{c|}{\small $97.43$} \\ \cline{2-7} 
\multirow{-2}{*}{2} & \cellcolor[HTML]{FFF5E6}\small FedFwd (FF) & \cellcolor[HTML]{FFF5E6}\small 0.64M & \multicolumn{2}{c|}{\cellcolor[HTML]{FFF5E6}\small $96.63$} & \multicolumn{2}{c|}{\cellcolor[HTML]{FFF5E6}\small $92.47$} \\ \hline
 & \small FedAvg (BP) & \small 0.89M & \multicolumn{2}{c|}{\small $98.25$} & \multicolumn{2}{c|}{\small $97.46$} \\ \cline{2-7} 
\multirow{-2}{*}{3} & \cellcolor[HTML]{FFF5E6}\small FedFwd (FF) & \cellcolor[HTML]{FFF5E6}\small 0.89M & \multicolumn{2}{c|}{\cellcolor[HTML]{FFF5E6}\small $96.78$} & \multicolumn{2}{c|}{\cellcolor[HTML]{FFF5E6}\small $92.57$} \\ \hline
\end{tabular}%
}
\caption{The comparison results between \fedfwd (FF) and \fedavg (BP) on MNIST dataset.}
\label{tab:expBPvsFF}
\end{table}

We compare \fedfwd with \fedavg for both i.i.d. and non-i.i.d. data distributions on two neural networks of different sizes.
The results are presented in Table \ref{tab:expBPvsFF}.
Firstly, we find that \fedfwd achieves a performance level that is on a par with \fedavg for the i.i.d. setting.
Specifically, we observe for \fedfwd approximately $1.5\%$ performance degradation in terms of test accuracy.
However, the gap increases in the non-i.i.d. setting.
For example, \fedavg achieves $97.46\%$ test accuracy on the 3-layer model, while \fedfwd achieves $92.57\%$ test accuracy.

This suggests that \fedfwd is more sensitive to changes in the data distribution than \fedavg.
Figure \ref{fig:learnable} provides empirical evidence to support the claim.
\fedfwd is a greedy approach that optimizes an objective function for each layer.
This makes \fedfwd more dependent on its data than \fedavg. 
Consequently, \fedfwd has slower convergence and greater variation.

\begin{figure}[h]
	\centering
	\subfloat[\small Hidden Layer = 2]{{\includegraphics[width=0.24\textwidth]{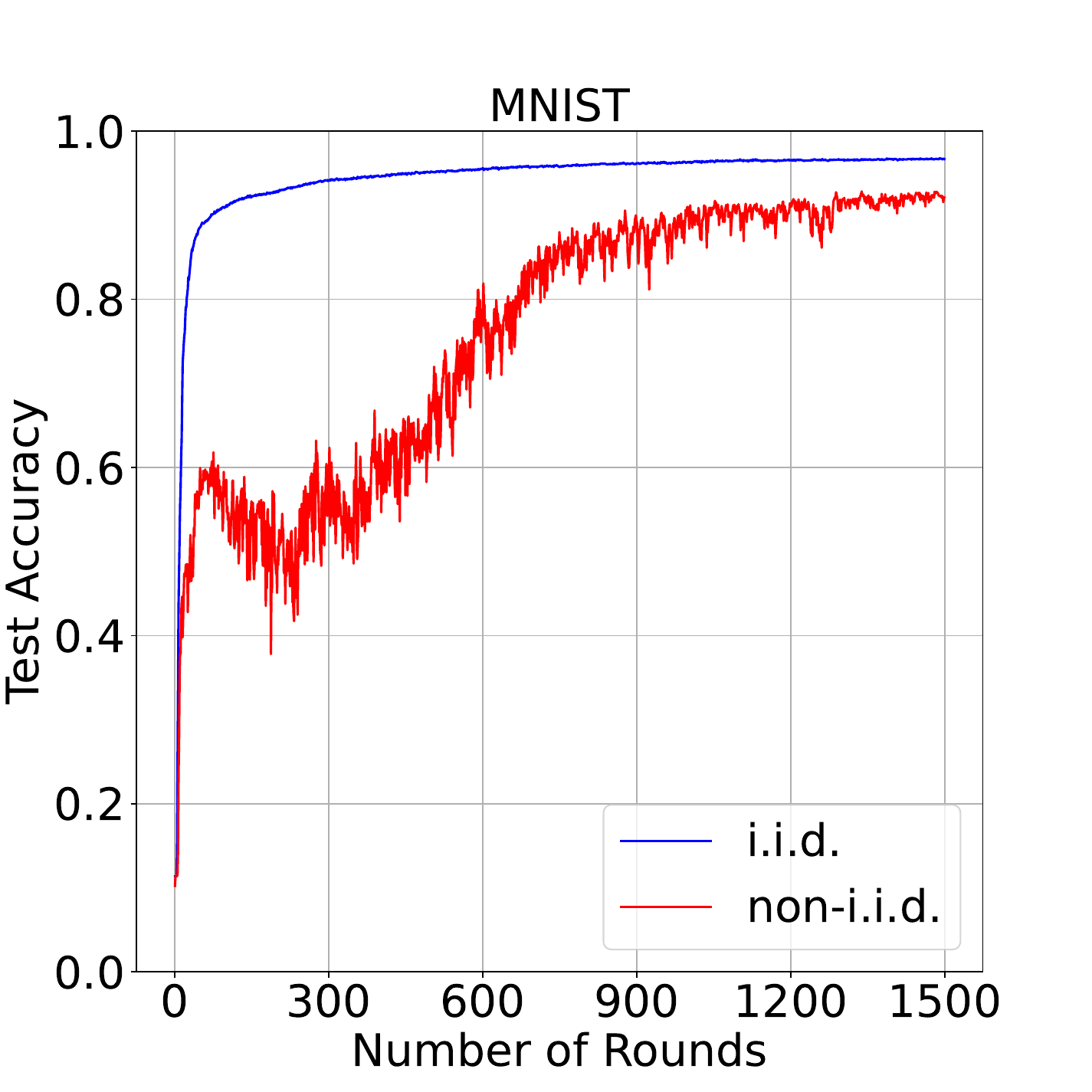} }}
 \subfloat[\small Hidden Layer = 3]{{\includegraphics[width=0.24\textwidth]{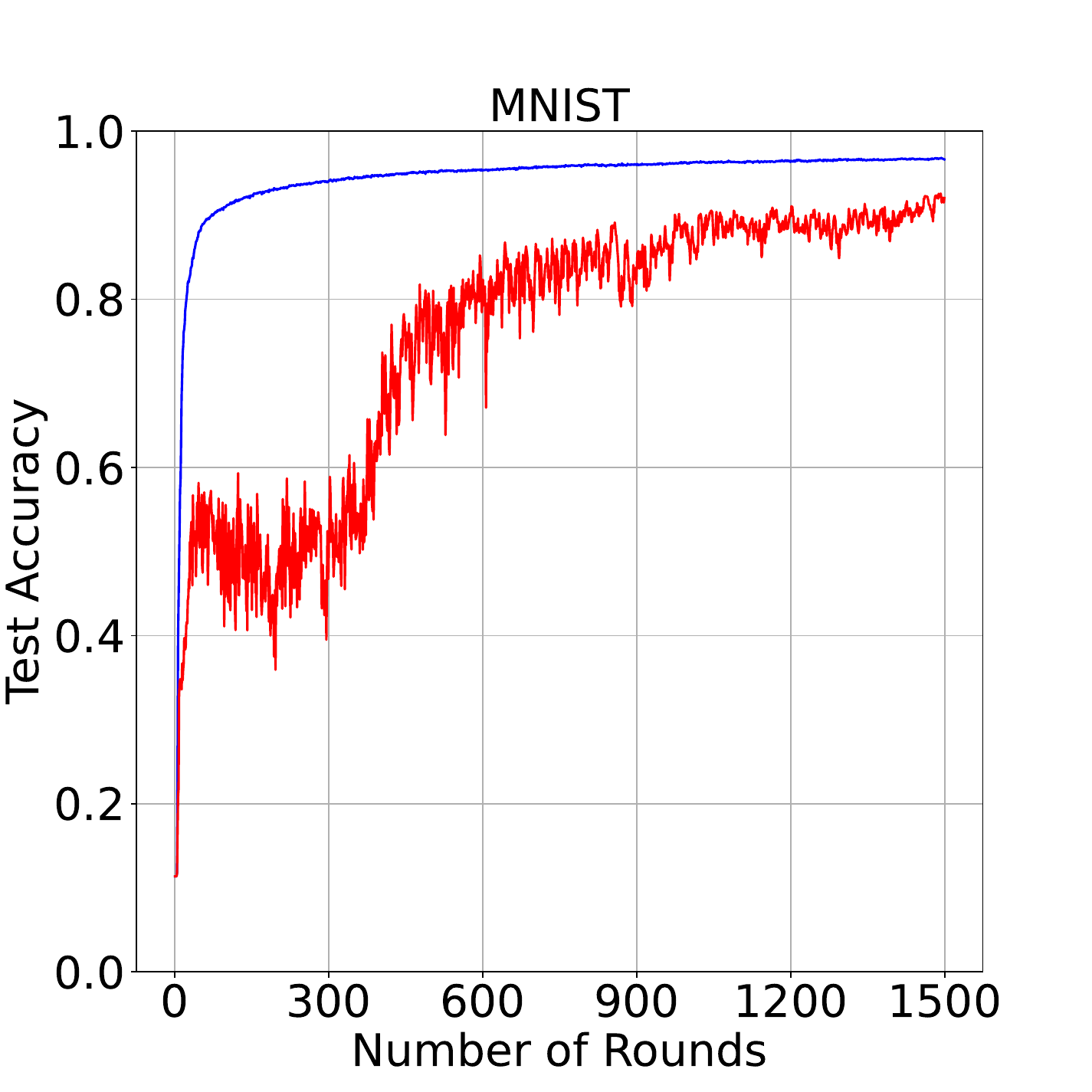} }}
 
 \caption{\small
		Training Stability of \fedfwd.
	}
	\label{fig:stable_learning}
\end{figure}

\begin{table*}[t]
\begin{minipage}{\linewidth}
  \begin{minipage}[b]{0.69\linewidth}
    \centering
    \resizebox{\textwidth}{!}{
    \begin{tabular}{c|c|ccccccccc}
    \hline
    \rowcolor[HTML]{EFEFEF} 
    \cellcolor[HTML]{EFEFEF} & \cellcolor[HTML]{EFEFEF} & \multicolumn{9}{c}{\cellcolor[HTML]{EFEFEF}\small Mini-Batch Size} \\ \cline{3-11} 
    \rowcolor[HTML]{EFEFEF} 
    \multirow{-2}{*}{\cellcolor[HTML]{EFEFEF}\begin{tabular}[c]{@{}c@{}}\small Dataset\\ \small (Size)\end{tabular}} & \multirow{-2}{*}{\cellcolor[HTML]{EFEFEF}\small Algorithm} & \multicolumn{1}{c|}{\cellcolor[HTML]{EFEFEF}\small 1} & \multicolumn{1}{c|}{\cellcolor[HTML]{EFEFEF}\small 4} & \multicolumn{1}{c|}{\cellcolor[HTML]{EFEFEF}\small 16} & \multicolumn{1}{c|}{\cellcolor[HTML]{EFEFEF}\small 64} & \multicolumn{1}{c|}{\cellcolor[HTML]{EFEFEF}\small 128} & \multicolumn{1}{c|}{\cellcolor[HTML]{EFEFEF}\small 256} & \multicolumn{1}{c|}{\cellcolor[HTML]{EFEFEF}\small 512} & \multicolumn{1}{c|}{\cellcolor[HTML]{EFEFEF}\small 1024} & \small 2048 \\ \hline
     & \small FedAvg (BP) & \multicolumn{1}{c|}{\small 10.92} & \multicolumn{1}{c|}{\small 4.09} & \multicolumn{1}{c|}{\small 3.01} & \multicolumn{1}{c|}{\small \textbf{2.20}} & \multicolumn{1}{c|}{\small 2.47} & \multicolumn{1}{c|}{\small 2.45} & \multicolumn{1}{c|}{\small 2.47} & \multicolumn{1}{c|}{\small 2.45} & \small 2.48 \\ \cline{2-11} 
    \multirow{-2}{*}{\begin{tabular}[c]{@{}c@{}}\small CIFAR-10\\ \small (50000)\end{tabular}} & \cellcolor[HTML]{FFF5E6}\small FedFwd (FF) & \multicolumn{1}{c|}{\cellcolor[HTML]{FFF5E6}\small 71.85} & \multicolumn{1}{c|}{\cellcolor[HTML]{FFF5E6}\small 20.92} & \multicolumn{1}{c|}{\cellcolor[HTML]{FFF5E6}\small 7.33} & \multicolumn{1}{c|}{\cellcolor[HTML]{FFF5E6}\small 3.77} & \multicolumn{1}{c|}{\cellcolor[HTML]{FFF5E6}\small 3.15} & \multicolumn{1}{c|}{\cellcolor[HTML]{FFF5E6}\small 2.88} & \multicolumn{1}{c|}{\cellcolor[HTML]{FFF5E6}\small 2.73} & \multicolumn{1}{c|}{\cellcolor[HTML]{FFF5E6}\small \textbf{2.24}} & \cellcolor[HTML]{FFF5E6} \small 2.64\\ \hline
     & \small FedAvg (BP) &\multicolumn{1}{c|}{\small 9.97} & \multicolumn{1}{c|}{\small 3.61} & \multicolumn{1}{c|}{\small 2.22} & \multicolumn{1}{c|}{\small 1.71} &\multicolumn{1}{c|}{\small 1.61} & \multicolumn{1}{c|}{\small 1.55} & \multicolumn{1}{c|}{\small \textbf{1.36}} & \multicolumn{1}{c|}{\small 1.53} & \multicolumn{1}{c}{\small 1.34}  \\ \cline{2-11} 
    \multirow{-2}{*}{\begin{tabular}[c]{@{}c@{}}\small MNIST\\ \small (60000)\end{tabular}} & \cellcolor[HTML]{FFF5E6}\small FedFwd (FF) & \multicolumn{1}{c|}{\cellcolor[HTML]{FFF5E6}\small 74.80} & \multicolumn{1}{c|}{\cellcolor[HTML]{FFF5E6}\small 23.11} & \multicolumn{1}{c|}{\cellcolor[HTML]{FFF5E6}\small 7.45} & \multicolumn{1}{c|}{\cellcolor[HTML]{FFF5E6}\small 2.64} & \multicolumn{1}{c|}{\cellcolor[HTML]{FFF5E6}\small 2.38} & \multicolumn{1}{c|}{\cellcolor[HTML]{FFF5E6}\small 1.74} & \multicolumn{1}{c|}{\cellcolor[HTML]{FFF5E6}\small 1.83} & \multicolumn{1}{c|}{\cellcolor[HTML]{FFF5E6}\small 1.77} & \cellcolor[HTML]{FFF5E6} \small \textbf{1.72}\\ \hline
    \end{tabular}}
  \captionof{table}{\small \textbf{Comparison in training time (Wall Clock Time) between \fedfwd and \fedavg.} 
    We compare the training time of \fedfwd and \fedavg under i.i.d. settings, analyzing how this varies with different mini-batch sizes. We measure the time it takes to train one global round. The results show that with a batch size of 1, the training time of \fedfwd is roughly seven times more than that of \fedavg.
    However, as the batch size increases, this difference decreases significantly.
    }
  \label{tab:speedff}
\end{minipage}
\hfill
\begin{minipage}[b]{0.3\linewidth}
    \centering
    \includegraphics[width=0.98\textwidth ]{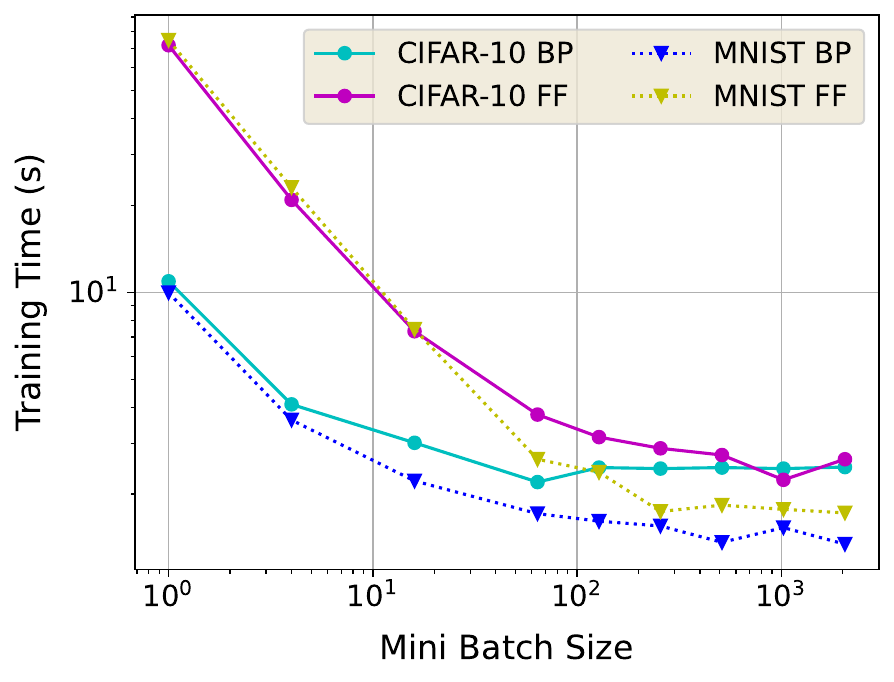}
    \captionof{figure}{Comparison in Speed}
    \label{fig:speed}
  \end{minipage}
\end{minipage}
\end{table*}

Next, we compare the training speeds of \fedfwd and \fedavg with respect to mini-batch sizes, as shown in Table \ref{tab:speedff} and Figure \ref{fig:speed}.
The results show that with a batch size of 1, the training speed of \fedfwd is roughly seven times slower than that of \fedavg.
However, as the batch size increase, this difference decrease significantly.
Since \fedavg is library-optimized while \fedfwd is not, a direct comparison of their wall clock times may not provide a complete picture. Nevertheless, this result clearly demonstrates the potential of \fedfwd.
It shows comparable, and in some cases superior, training times to \fedavg as batch sizes increase.
Looking ahead, we anticipate that with further improvements in both software and hardware optimization, \fedfwd will match or even exceed \fedavg's learning speeds, even with smaller mini-batches, ultimately reducing the time burden on clients in a federated setting.
\section{Conclusion}
In this work, we propose \fedfwd, a new federated optimization algorithm that trains local models using FF before uplink communication.
This approach could be more memory- and computationally- efficient than BP-based FL algorithms.
We evaluate \fedfwd on two datasets MNIST and CIFAR-10.
Our results indicate that \fedfwd achieves comparable test accuracy and training time to \fedavg, a standard BP-based FL algorithm.
Additionally, \fedfwd demonstrates potential for reducing training time and improving learning stability on a new objective function.
We believe that  \fedfwd will be a promising new approach to FL and inspire further research on more efficient FL algorithms.

\section{Acknowledgement}
This work was partly supported by Institute of Information \& communications Technology Planning \& Evaluation (IITP) grant funded by the Korea government (MSIT) (No.2019-0-01906, Artificial Intelligence Graduate School Program (POSTECH)) and National Research Foundation of Korea (NRF) grant funded by the Korea government (MSIT) (RS-2023-00210466).

\nocite{langley00}

\bibliography{ref}
\bibliographystyle{icml2023}

%%%%%%%%%%%%%%%%%%%%%%%%%%%%%%%%%%%%%%%%%%%%%%%%%%%%%%%%%%%%%%%%%%%%%%%%%%%%%%%
%%%%%%%%%%%%%%%%%%%%%%%%%%%%%%%%%%%%%%%%%%%%%%%%%%%%%%%%%%%%%%%%%%%%%%%%%%%%%%%
% APPENDIX
%%%%%%%%%%%%%%%%%%%%%%%%%%%%%%%%%%%%%%%%%%%%%%%%%%%%%%%%%%%%%%%%%%%%%%%%%%%%%%%
%%%%%%%%%%%%%%%%%%%%%%%%%%%%%%%%%%%%%%%%%%%%%%%%%%%%%%%%%%%%%%%%%%%%%%%%%%%%%%%
\newpage
\appendix
\onecolumn

\section{Models}
\label{app:models}
\subsection{Models}
\subsubsection{How does the model train?}
The \fedfwd algorithm fundamentally flattens all images and creates positive and negative data.
Upon entering the first hidden layer, activity is generated within each unit for the positive and negative data. 
By summing up the squares of the activity in the first hidden layer, the goodness of the corresponding layer is determined. 
In the corresponding layer, the goodness of positive data is optimized to be greater than a specific threshold, $\theta$, and the goodness of negative data is optimized to be less than a specific threshold. 
To efficiently learn new features for each layer when stacking multiple layers, the value of activity is not directly passed to the next layer; 
only the directional component is transmitted. 
In other words, all activities are normalized before forwarding to the next layer, only conveying orientation not size. 
This process is repeated for each layer, and the entire model is learned in a greedy layer-wise manner so that each layer effectively discovers new features through goodness.

\begin{figure*}[h]
	\centering
	\includegraphics[width=0.48\textwidth]{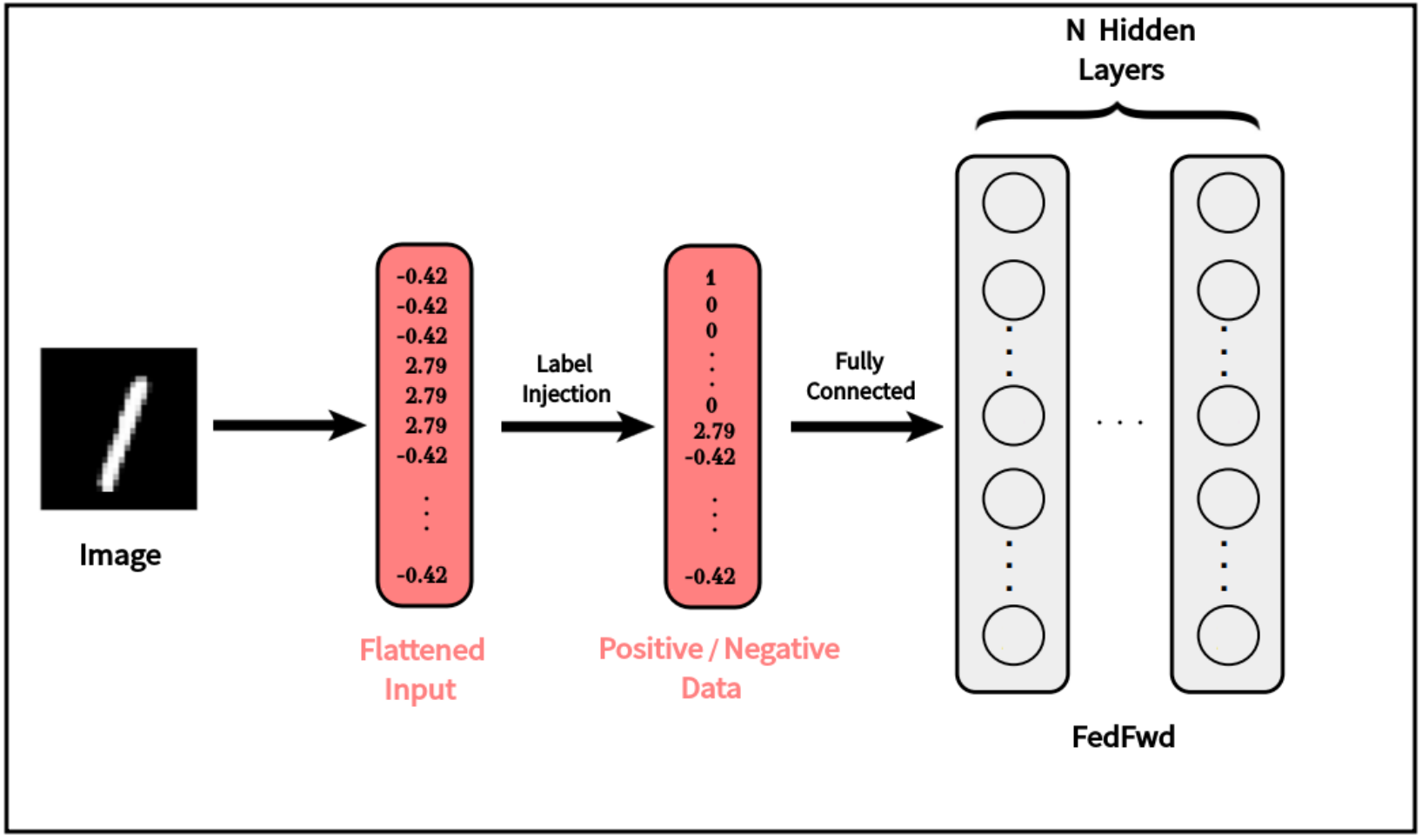}
	\caption{\small
		Model Architecture of \fedfwd.
	}
	\vspace{-1.5em}
	\label{fig:Model}
\end{figure*}

\subsubsection{How Does It Predict?}
During the prediction phase, a certain portion of the pixels in the flattened image is replaced by the one-hot vector of the correct label, similar to the training process. 
However, unlike the training phase, the prediction phase involves selecting the correct answer based on the total goodness value for each layer, rather than calculating and optimizing the goodness per layer.
Furthermore, during prediction, images corresponding to all labels are generated and fed into the \fedfwd model, as opposed to just one label during training.
For example, in the MNIST dataset, the one-hot vector of all labels is inserted at the beginning of the image, and the sum of the goodness values across all layers is compared when the resulting ten labeled images are input into the model. 
The correct answer is then determined by selecting the label with the maximum goodness value across all layers.

\begin{figure*}[h]
	\centering
	\includegraphics[width=0.8\textwidth]{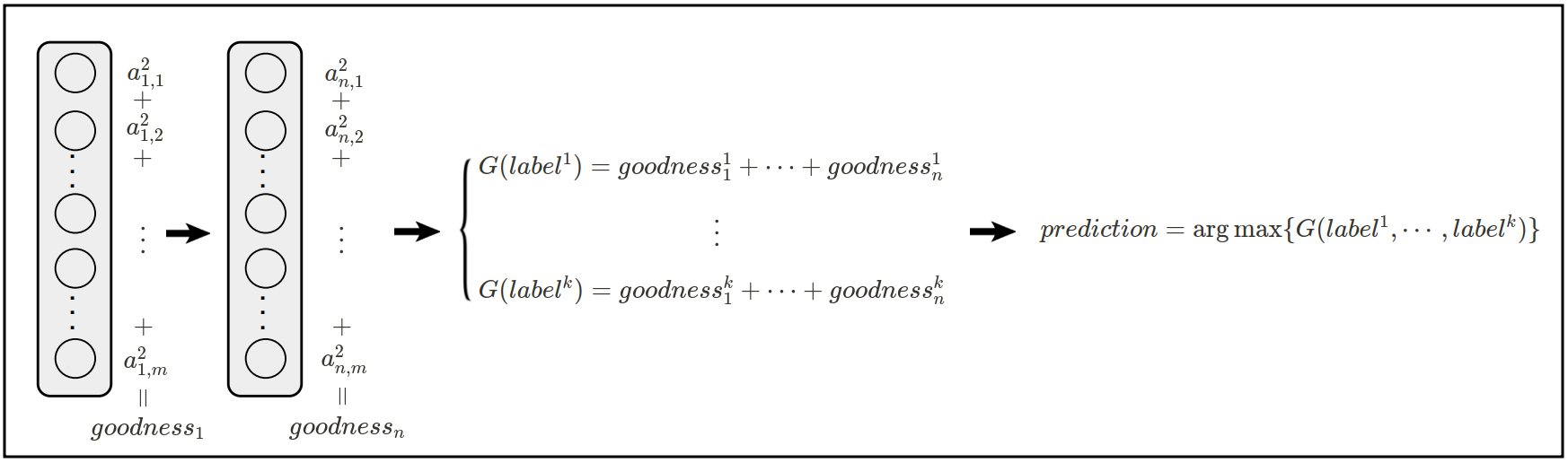}
	\caption{\small
		How Model Predicts in \fedfwd.
	}
	\vspace{-1.5em}
	\label{fig:PredModel}
\end{figure*}

\section{Additional Experiments}
\label{app:overallexp}

\subsection{More on Layer Depth \& Layer Size \& Heterogeneity in CIFAR-10 Dataset}
\label{app:addexp}
In this study, we conducted a series of experiments applying the \fedfwd algorithm in federated learning environments, utilizing the CIFAR-10 \citep{krizhevsky2009learning} dataset to exhibit its feasible performance capabilities. 
We undertook an analysis of test accuracy, fluctuating both the size and depth of hidden layers for a comprehensive overview. 
The experiments spanned over 1500 epochs, and our results suggest that while performance may be slightly underwhelming, \fedfwd enables adequate learning. 
For better visualization and understanding of the learning graph, we represented the y-axis based on a 50\% test accuracy benchmark instead of the usual 100\%.

\begin{figure*}[h]
    \centering
    \subfloat[Hidden Layer = 2]{{\includegraphics[width=0.33\textwidth]{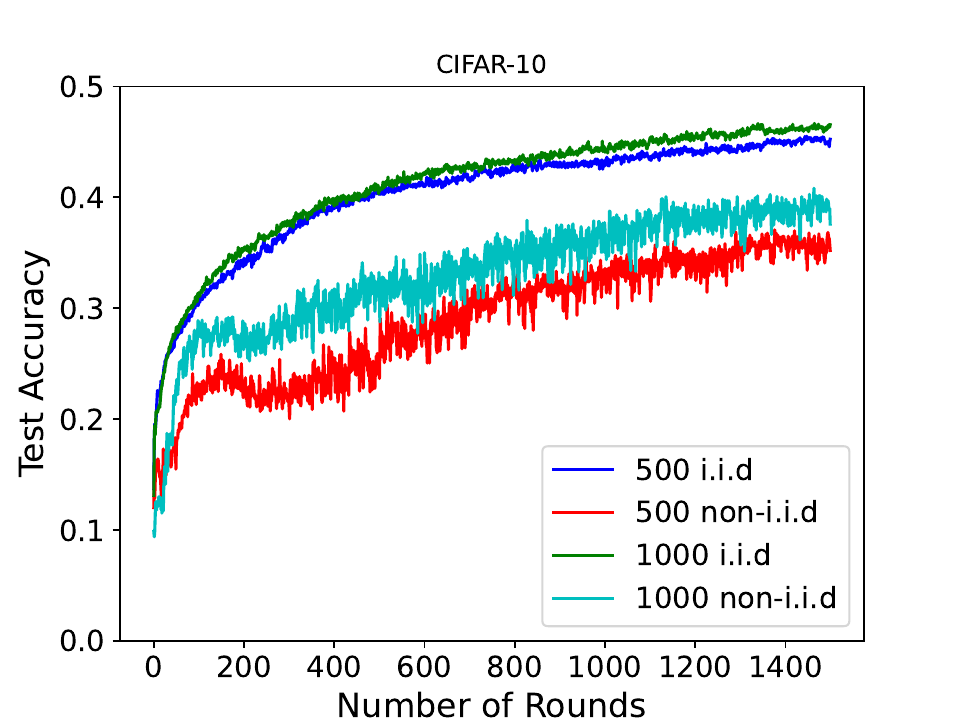} }}
    \subfloat[Hidden Layer = 3]{{\includegraphics[width=0.33\textwidth]{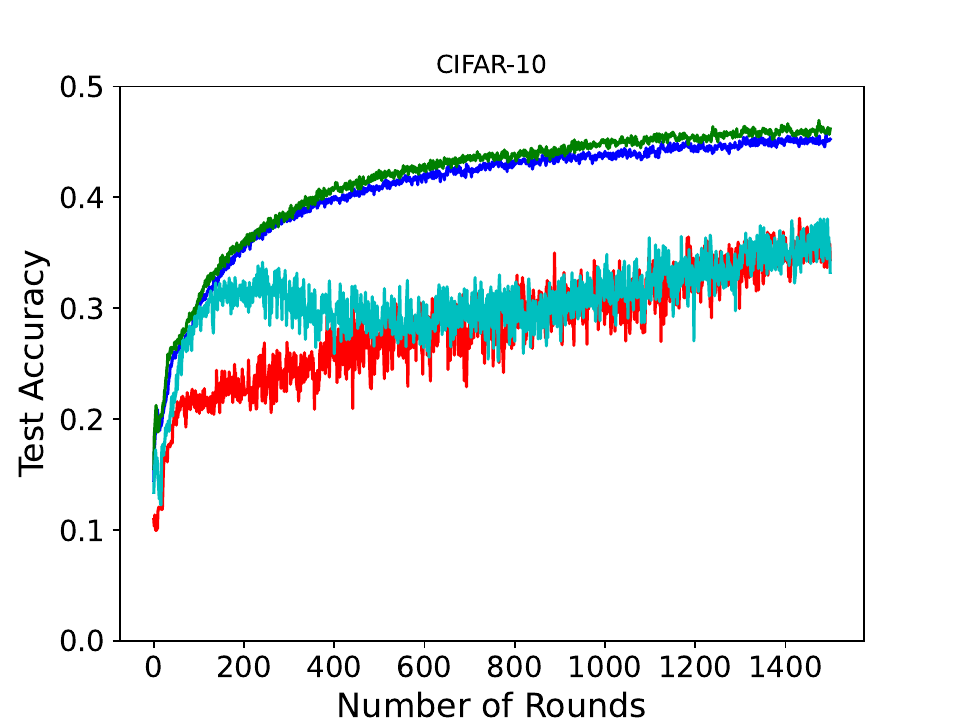} }}
    \subfloat[Hidden Layer = 4]{{\includegraphics[width=0.33\textwidth]{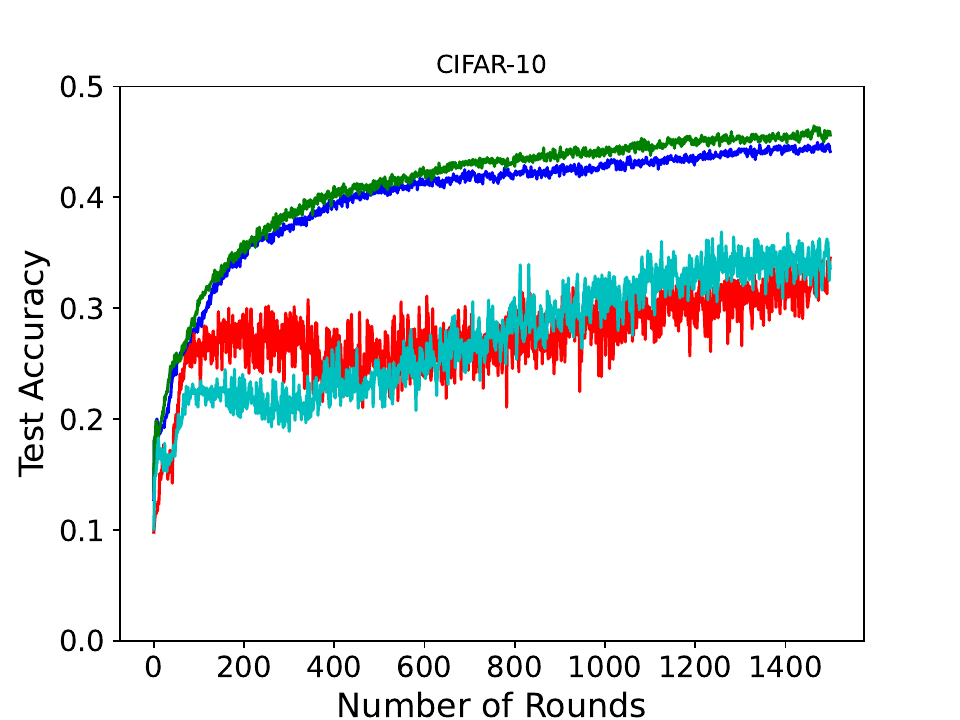} }}
    \captionsetup{width=.6\textwidth}
    \caption{
        \small \textbf{Results of \fedfwd varying layer depth, size, and heterogeneity using CIFAR-10.} 
        \small Specifically, we adjusted the depth of the hidden layer to 2, 3, and 4, and the size to 500 and 1000, analyzing test accuracy in each case. 
        These tests were further categorized into i.i.d. and non-i.i.d. settings, and were all conducted over 1500 epochs. 
        The results reveal that the \fedfwd algorithm is compatible with and can learn from the CIFAR-10 dataset.
    }
\end{figure*}

\begin{table}[h]
\centering
\resizebox{0.5\columnwidth}{!}{%
\begin{tabular}{|c|c|c|cl|cl|}
\hline
\rowcolor[HTML]{EFEFEF} 
\small \# Layers & \begin{tabular}[c]{@{}c@{}}\small Learning\\\small Procedure\end{tabular} & \small \# Params & \multicolumn{2}{c|}{\cellcolor[HTML]{EFEFEF}\begin{tabular}[c]{@{}c@{}}\small I.I.D.\\\small  Test Acc (\%)\end{tabular}} & \multicolumn{2}{c|}{\cellcolor[HTML]{EFEFEF}\begin{tabular}[c]{@{}c@{}}\small Non-I.I.D.\\\small  Test Acc (\%)\end{tabular}} \\ \hline
 & \small FedAvg (BP) &\small  1.79M & \multicolumn{2}{c|}{\small $58.86$} & \multicolumn{2}{c|}{\small $50.67$} \\ \cline{2-7} 
\multirow{-2}{*}{2} & \cellcolor[HTML]{FFF5E6}\small FedFwd (FF) & \cellcolor[HTML]{FFF5E6}\small 1.78M & \multicolumn{2}{c|}{\cellcolor[HTML]{FFF5E6}\small $45.51$} & \multicolumn{2}{c|}{\cellcolor[HTML]{FFF5E6}\small $37.06$} \\ \hline
 & \small FedAvg (BP) & \small 2.04M & \multicolumn{2}{c|}{\small $59.91$} & \multicolumn{2}{c|}{\small $52.00$} \\ \cline{2-7} 
\multirow{-2}{*}{3} & \cellcolor[HTML]{FFF5E6}\small FedFwd (FF) & \cellcolor[HTML]{FFF5E6}\small 2.03M & \multicolumn{2}{c|}{\cellcolor[HTML]{FFF5E6}\small $45.76$} & \multicolumn{2}{c|}{\cellcolor[HTML]{FFF5E6}\small $38.10$} \\ \hline
\end{tabular}%
}
\caption{The comparison results between \fedfwd (FF) and \fedavg (BP) on CIFAR-10 dataset.}
\label{tab:expBPvsFFCIFAR}
\end{table}

As presented in Table \ref{tab:expBPvsFFCIFAR}, we compared the performance of both algorithms on the CIFAR-10 dataset using models with nearly the same number of parameters, with a difference of only 1\%. 
Neither \fedavg (BP) nor \fedfwd (FF) exhibit high performance or demonstrate successful learning. 
This may be due to the simplistic approach of flattening the complex image information of CIFAR-10 to learn features. 
Nonetheless, when comparing the results, the \fedavg method displays significantly higher performance than \fedfwd across all layers in both i.i.d. and non-i.i.d. settings (approximately 13-14\%). 
This indicates that the current \fedfwd approach requires further investigation and development in terms of model architecture, objective function, and other yet-to-be-discovered performance enhancement methods. 
In response, we conducted additional research on modifying the objective function, which can be found in Appendix \ref{app:obj}.

\subsection{Additional Results on the Effect of Objectives in the \fedfwd Algorithm in Federated Learning}
\label{app:obj}

Below, we present experiments aimed at improving the learning stability of the \fedfwd algorithm. 
To achieve this, we adapted the \Symba \citep{lee2023symba} algorithm for a federated learning setting. 
The modified algorithm enhances convergence speed and performance by properly balancing positive and negative losses, addressing the issue of conflicting convergence directions for positive and negative samples in the original FF algorithm. 
By testing the \Symba approach in a federated setting, particularly in non-i.i.d. settings, we were able to achieve more stable learning and faster convergence as well as slight performance improvements.
Through this, we demonstrate that if the initially proposed \fedfwd algorithm is further developed, it can not only exhibit advanced learning stability and performance but also be improved potentially in numerous unexplored research areas. 
Therefore, \textbf{it merits further investigation.}

\FloatBarrier
\begin{figure*}[h]
    \centering
    \subfloat[Hidden Layer = 2]{{\includegraphics[width=0.3\textwidth ]{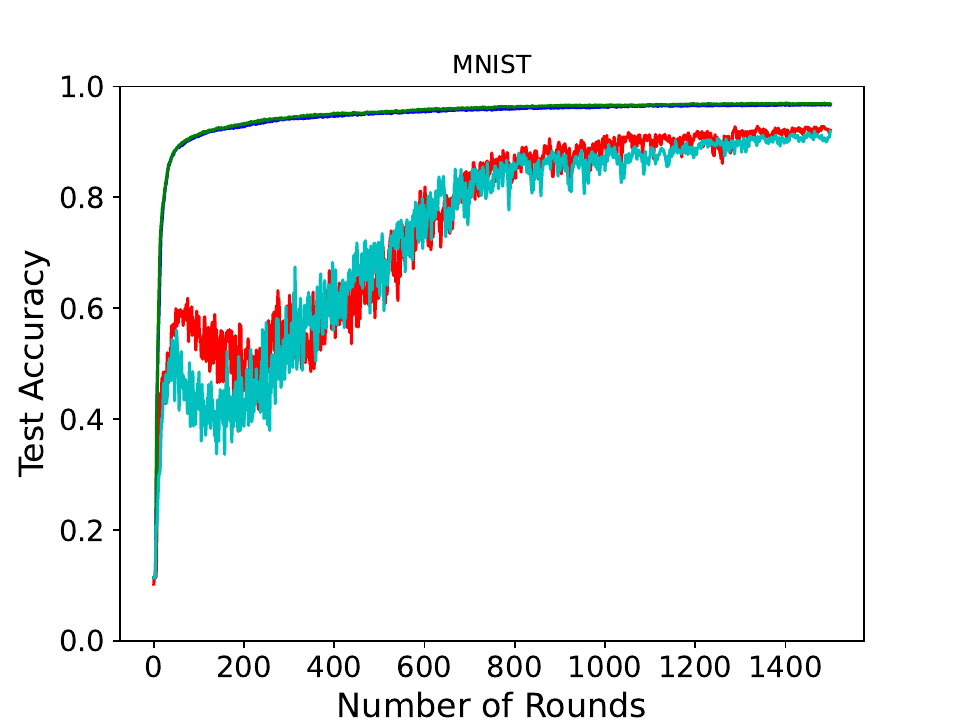} }}
    \subfloat[Hidden Layer = 3]{{\includegraphics[width=0.3\textwidth ]{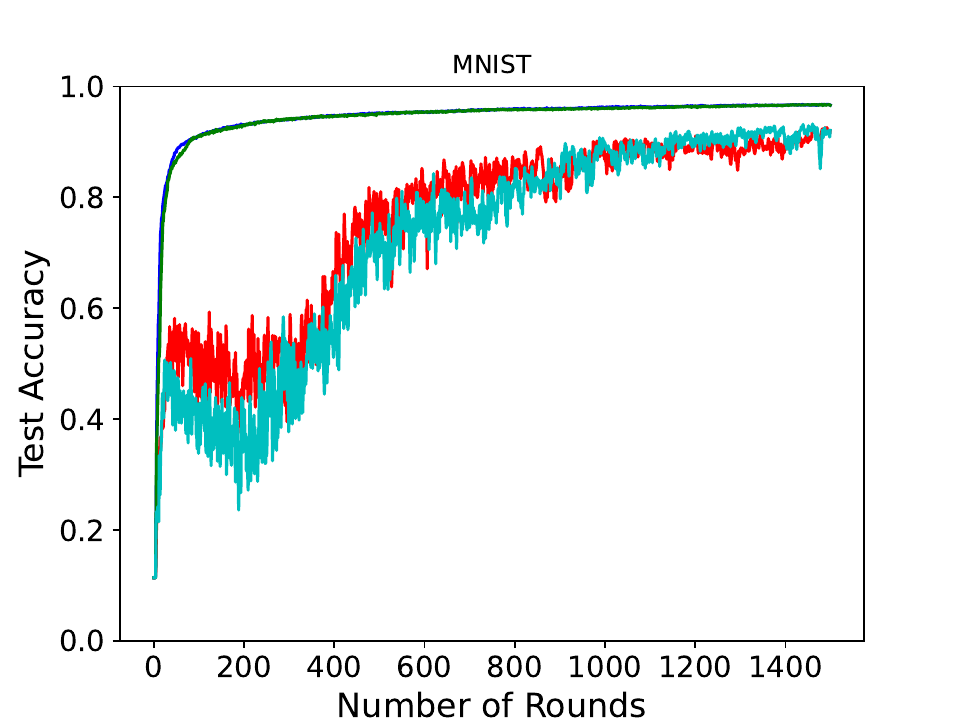} }}
    \subfloat[Hidden Layer = 4]{{\includegraphics[width=0.3\textwidth ]{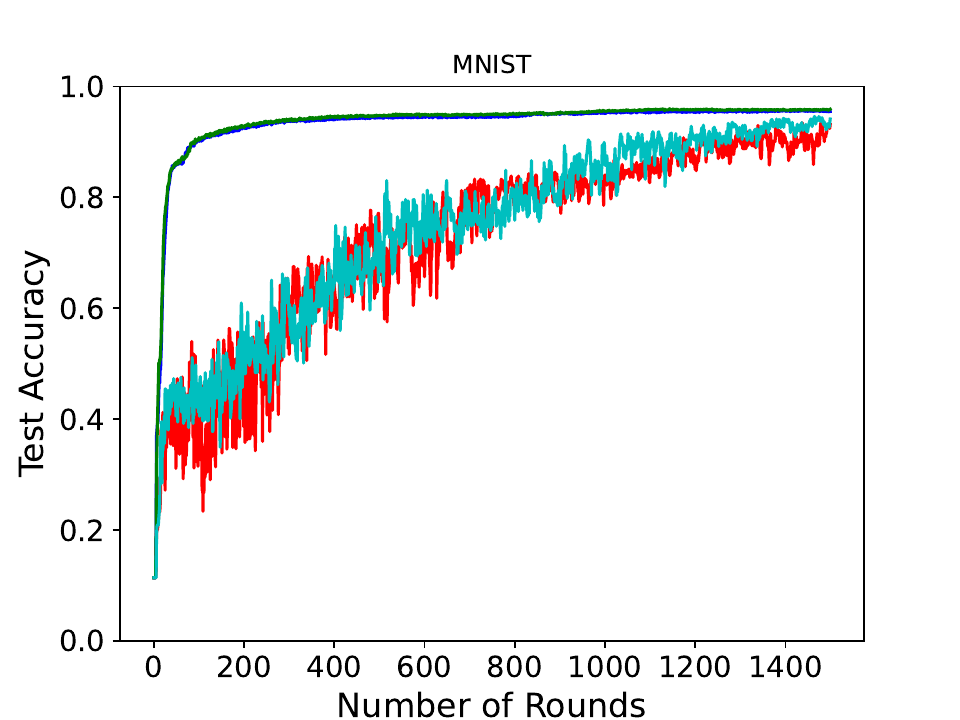} }}
    \hfill
    \subfloat[Hidden Layer = 2]{{\includegraphics[width=0.3\textwidth ]{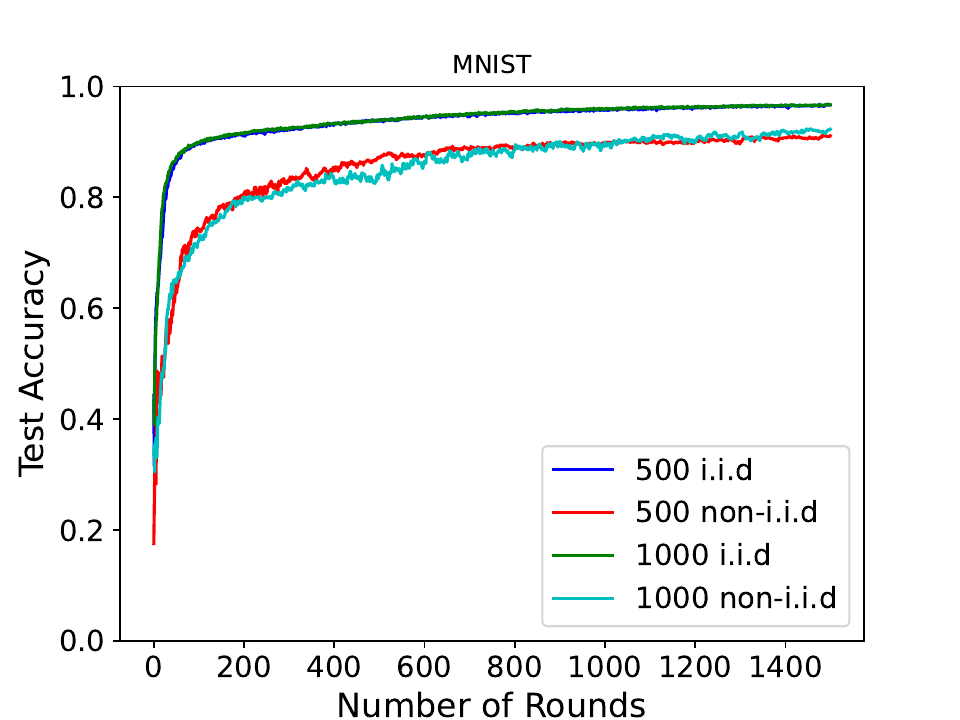} }}
    \subfloat[Hidden Layer = 3]{{\includegraphics[width=0.3\textwidth ]{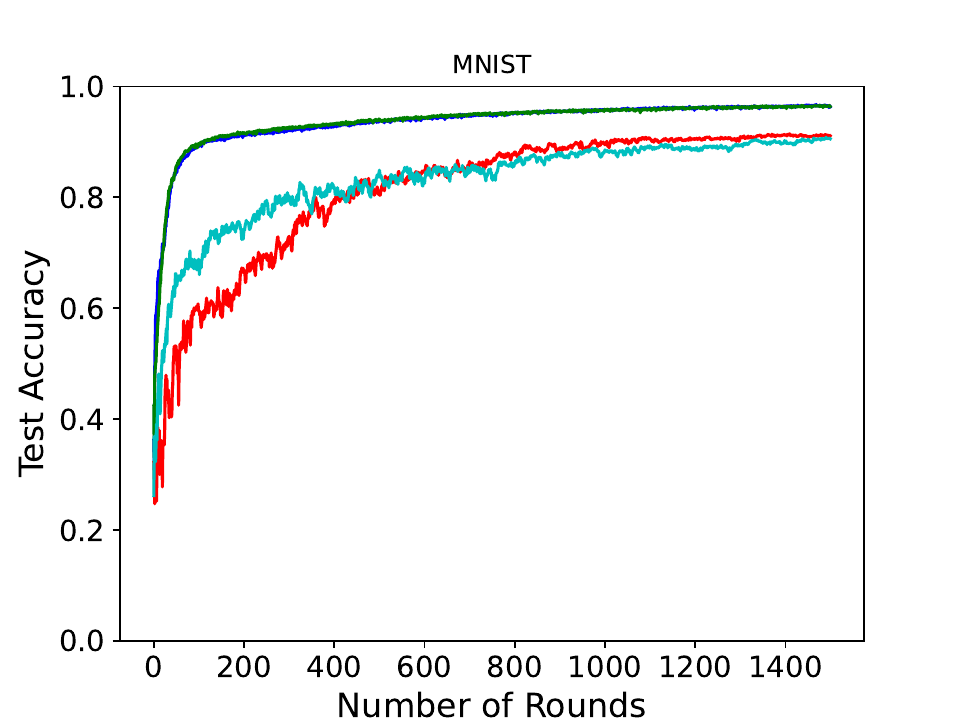} }}
    \subfloat[Hidden Layer = 4]{{\includegraphics[width=0.3\textwidth ]{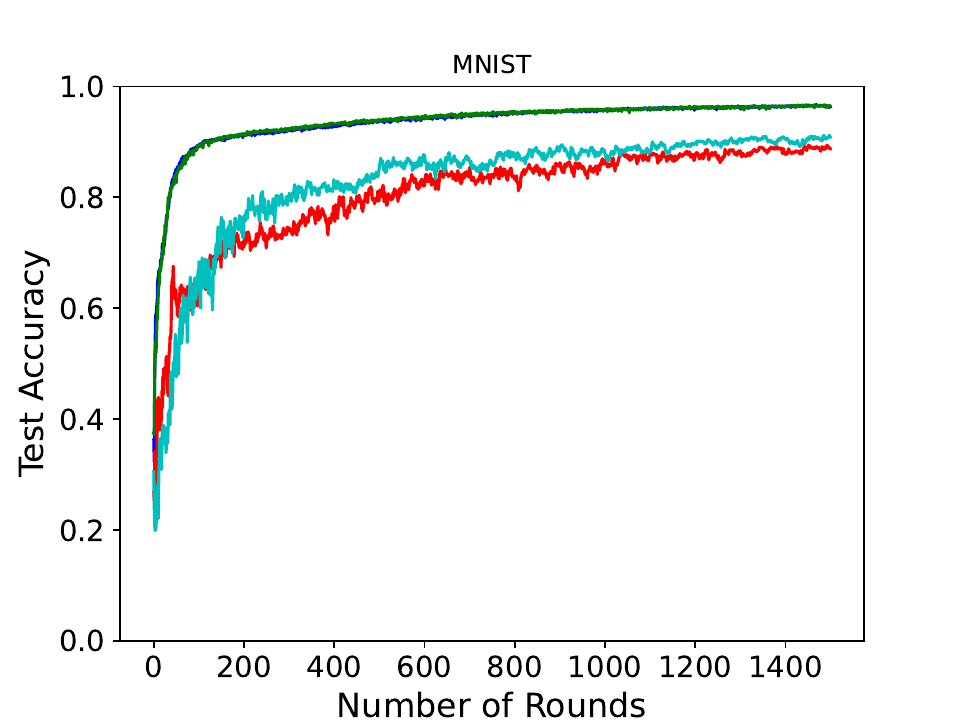} }}
    \caption{
        \small \textbf{Result of combining \fedfwd and \Symba on MNIST Dataset.}
        \small Our illustrations, namely figures (a), (b), and (c), depict the original \fedfwd algorithm, while figures (d), (e), and (f) represent the amalgamation of the \fedfwd and \Symba algorithms. 
        This amalgamated approach yields more stable learning results, exhibiting lower variances, and quicker convergence in comparison to the stand-alone \fedfwd algorithm.
    }

    \vspace{10mm}
    \centering
    \subfloat[Hidden Layer = 2]{{\includegraphics[width=0.3\textwidth ]{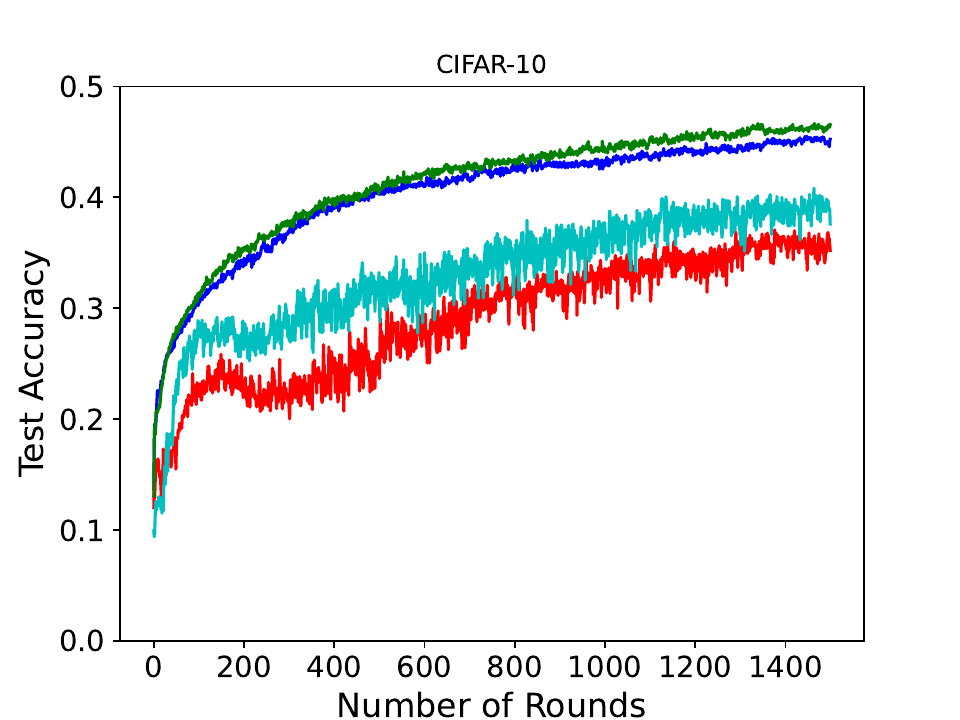} }}
    \subfloat[Hidden Layer = 3]{{\includegraphics[width=0.3\textwidth ]{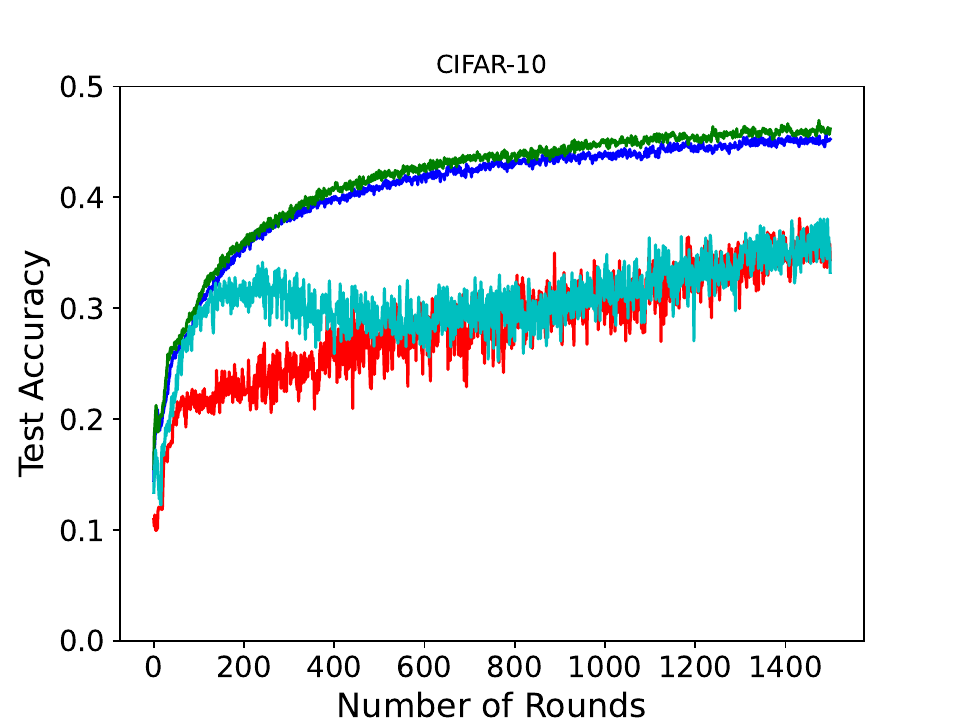} }}
    \subfloat[Hidden Layer = 4]{{\includegraphics[width=0.3\textwidth ]{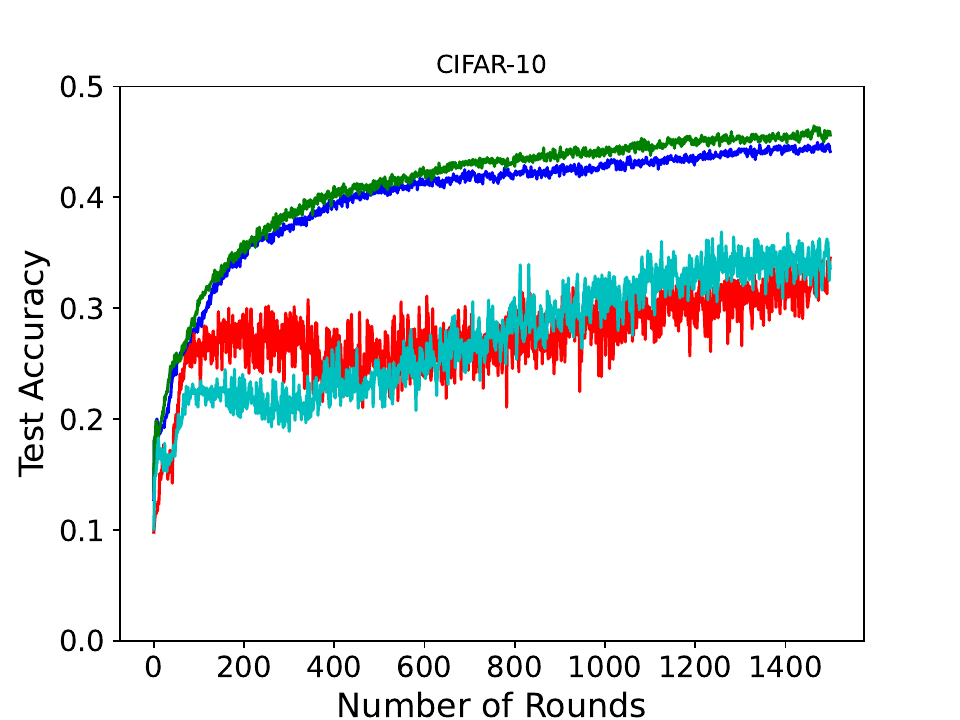} }}
    \hfill
    \subfloat[Hidden Layer = 2]{{\includegraphics[width=0.3\textwidth ]{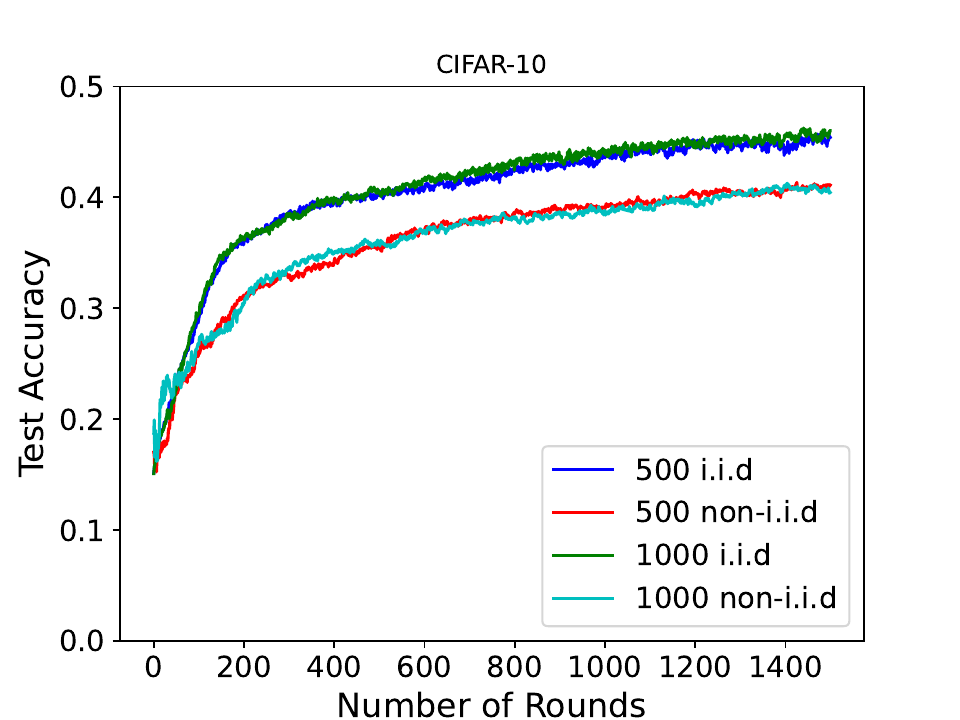} }}
    \subfloat[Hidden Layer = 3]{{\includegraphics[width=0.3\textwidth ]{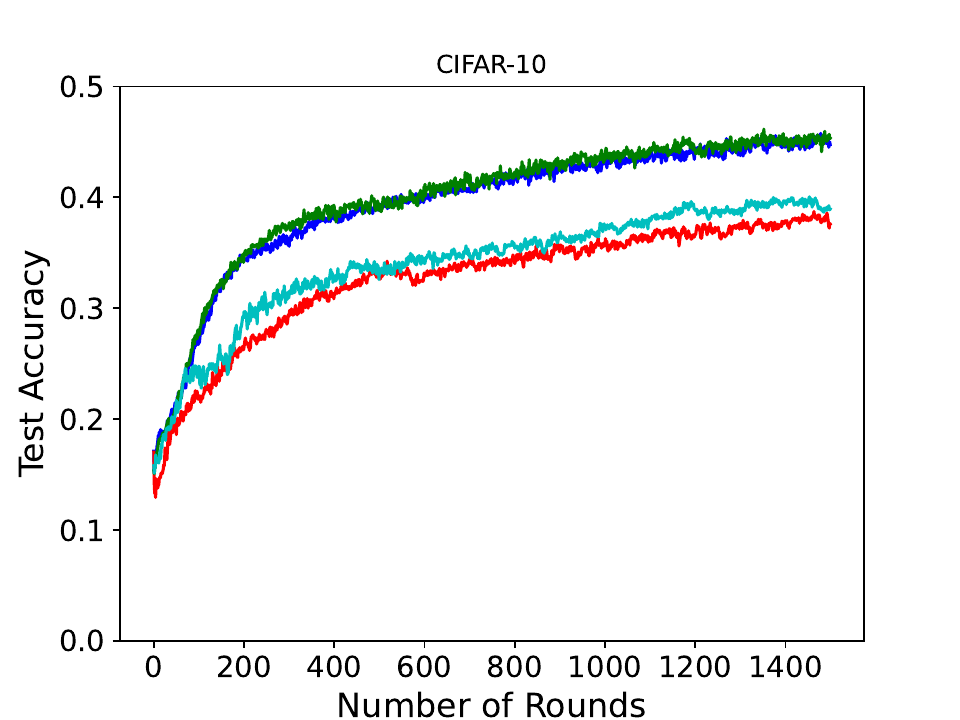} }}
    \subfloat[Hidden Layer = 4]{{\includegraphics[width=0.3\textwidth ]{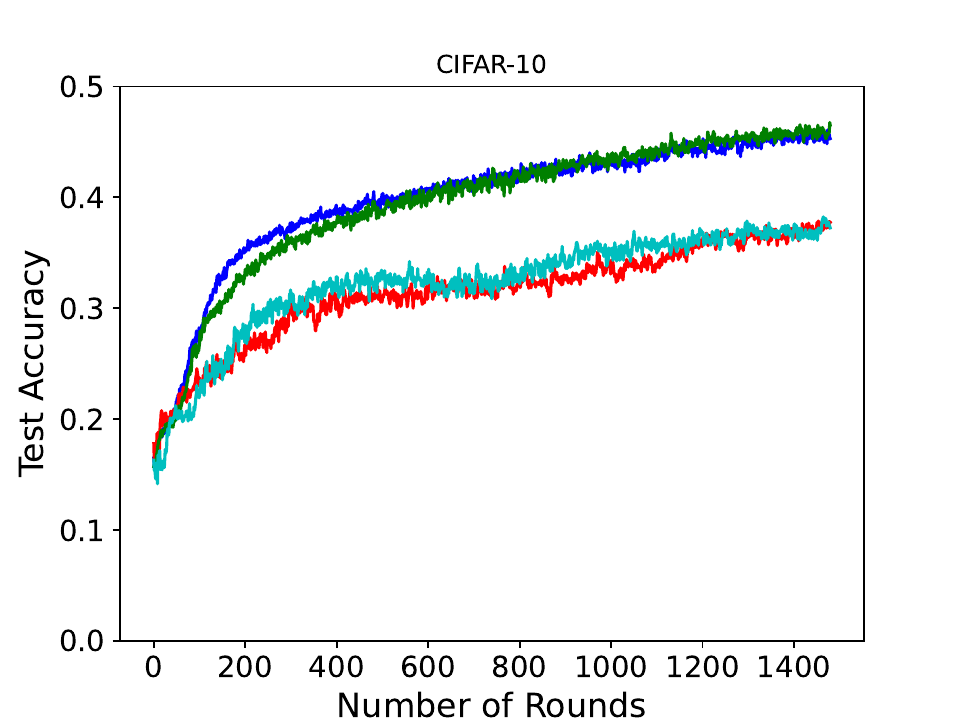} }}
    \caption{
        \small \textbf{Result of combining \fedfwd and \Symba on CIFAR-10 Dataset.}
        \small Figures (a), (b), and (c) denote the original \fedfwd algorithm, while figures (d), (e), and (f) depict the combination of the \fedfwd and \Symba algorithms. 
        To facilitate a clearer representation of the learning graph, we've calibrated the y-axis to reflect 50\% test accuracy instead of 100\%. 
        In testing with the CIFAR-10 dataset, we've observed slightly superior results in terms of convergence rate, performance, and learning stability, comparable to what was observed with original \fedfwd.
    }
\end{figure*}
\FloatBarrier

\end{document}